\documentclass[10pt,twocolumn,letterpaper]{article}

\usepackage{wacv}              

\definecolor{wacvblue}{rgb}{0.21,0.49,0.74}
\usepackage[pagebackref,breaklinks,colorlinks,allcolors=wacvblue]{hyperref}

\usepackage[accsupp]{axessibility}
\usepackage{graphicx} 
\usepackage[table]{xcolor} 
\usepackage{amsmath}              
\usepackage{float}                
\usepackage{caption}              
\usepackage{algorithm}            
\usepackage{algpseudocode}        
\usepackage[utf8]{inputenc} 
\usepackage[T1]{fontenc}    
\usepackage{hyperref}       
\usepackage{url}            
\usepackage{booktabs}       
\usepackage{amsfonts}       
\usepackage{nicefrac}       
\usepackage{microtype}      
\usepackage{xcolor}         
\usepackage{graphicx}
\usepackage[most]{tcolorbox}
\usepackage{xcolor}
\usepackage{lipsum}
\usepackage[utf8]{inputenc}
\usepackage{makecell}
\usepackage{subcaption}
\def\wacvPaperID{3346} 
\def\confName{WACV}
\def\confYear{2026}

\makeatletter

\title{\LARGE \bfseries See, Think, Learn: A Self-Taught Multimodal Reasoner}

\author{
Sourabh Sharma$^{1}$ \quad
Sonam Gupta$^{2}$ \quad
Sadbhawna$^{1}$\\[4pt]
{\tt\small sourabh125ss@gmail.com \quad sonam.gupta7@ibm.com \quad sadbhawna.cse@mnit.ac.in}\\[4pt]
$^{1}$ Malaviya National Institute of Technology Jaipur \quad $^{2}$ IBM Research 
}

\begin{document}     
\maketitle

\begin{abstract}
Vision-Language Models (VLMs) have achieved remarkable progress in integrating visual perception with language understanding. However, effective multimodal reasoning requires both accurate \textbf{perception} and robust \textbf{reasoning}, and weakness in either limits the performance of VLMs. Prior efforts to enhance reasoning often depend on high-quality chain-of-thought (CoT) data, obtained via labor-intensive human annotations, costly proprietary models or self-training methods that overlook perception. To address these limitations, we propose a simple yet effective self-training framework called "See-Think-Learn" (STL). At its core, STL introduces a structured reasoning template that encourages the model to \textit{see before thinking}: first extracting visual attributes in textual form, then using them to guide reasoning. 
The framework jointly improves perception and reasoning by having the model generate and learn from its own structured rationales in a self-training loop. Furthermore, we augment the training data with negative rationales, i.e. explanations that justify why certain answer choices are incorrect, to enhance the model's ability to distinguish between correct and misleading responses. This fosters more discriminative and robust learning. Experiments across diverse domains show that STL consistently outperforms baselines trained directly only on answers or self-generated reasoning, while qualitative analysis confirms the high quality of its rationales. STL thus provides a cost-effective solution to enhance multimodal reasoning ability of VLMs. Our code is available at \href{https://github.com/srbhcs/see-think-learn}{https://github.com/srbhcs/see-think-learn}.
\end{abstract}
    
\section{Introduction}
Large Language Models (LLMs) have recently demonstrated remarkable advances in complex reasoning through the use of Chain-of-Thought (CoT) prompting \cite{letsthinkstepbystep, wei2022chain, yao2023tree, lyu2023faithful}. By explicitly encouraging models to generate intermediate reasoning steps before arriving at the final answer, CoT significantly boosts performance on a wide range of textual tasks \cite{plaat2024reasoning, chen2025towards, zhang2024llm, bandyopadhyay2025thinking}. Motivated by this success, recent works \cite{xu2411llava, wu2025boosting, mitra2024compositional} have attempted to extend CoT prompting to multimodal extensions of LLMs, termed as Visual Language Models (VLMs) \cite{liu2023improvedllava, liu2023llava, bai2308qwen, achiam2023gpt, team2023gemini, dong2024internlm}. 
However, reasoning in VLMs remains a fundamental challenge due to their limited capacity to jointly understand and reason over both visual and textual information.

\begin{figure}
\centering
  \includegraphics [width=.47\textwidth]{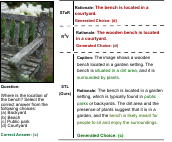}
\caption{\textbf{Comparison of reasoning generated by our ``See-Think-Learn” (STL) framework with STaR \cite{zelikman2022star} and R\textsuperscript{3}V \cite{cheng2025vision}}. STL produces more detailed and perceptually grounded rationales, whereas STaR and R\textsuperscript{3}V tend to overlook contextual cues and provide shorter, less comprehensive explanations.}
\label{fig:qual}
\end{figure}

A major bottleneck in training VLMs for CoT-style reasoning is the lack of high-quality supervision. Existing datasets \cite{lu2022learn, schwenk2022okvqa} are often limited to short answers with minimal or no explanatory rationales. For curating high-quality rationales, current research typically relies on labor-intensive human annotations (e.g., \cite{chen-etal-2024-m3cot, yue2024mmmu}) which are difficult to scale or on proprietary black-box models like GPT-4V \cite{openai2023gpt4} or Gemini \cite{yue2024mmmu} which are expensive.

To overcome these limitations, we propose a simple yet effective self-training framework called See-Think-Learn (STL). STL leverages the model's existing perception and reasoning capabilities to iteratively improve itself by generating and learning from its own rationales. A key question in this process is: \textit{What should the structure of these rationales be?}  
Effectively answering a visual question requires strong perception and reasoning.
To strengthen both components, STL introduces a structured rationale format grounded in the principle of “see before thinking”. Specifically, the model is prompted to (1) first describe visual elements from the image, (2) then reason about them in context, and (3) finally produce an answer. This structure mirrors the natural human cognitive process and guides the model in better organizing its internal reasoning.

Relying only on rationales from correct answers provides limited supervision, as the model observes only successful reasoning paths. This weakens performance on complex tasks where distinguishing correct from incorrect reasoning is critical. To address this, we introduce negative rationales (explanations of why certain answers are wrong) alongside positive ones. This mirrors reflective human learning, where understanding improves by analyzing both successes and mistakes. Negative rationales expose flawed reasoning patterns and highlight contrasts between correct and incorrect answers, helping the model avoid pitfalls such as hallucinations and unsupported inferences.

We operationalize this idea in our Self-Taught Multimodal Reasoner (STL), a scalable self-training framework where a VLM learns to generate both positive and negative rationales and iteratively improves through retraining on these structured annotations. As illustrated in Figure~\ref{fig:qual}, STL produces more detailed and accurate reasoning. Evaluations across commonsense, scientific, and language-based domains show that STL outperforms models trained on final answers alone and remains broadly comparable to models trained with human-annotated rationales (Table \ref{tab:m3cot_rationale_comparison}), despite some variation across datasets. These results highlight STL’s potential as a practical alternative to annotation-heavy approaches.

In summary, the contributions of this work are as follows:

\begin{enumerate}
    \item We introduce the "see-before-thinking" rationale structure, which explicitly separates perception and reasoning components within the generated rationale.
    \item We enhance training data with negative rationales, allowing the model to learn to distinguish between correct and incorrect reasoning paths.
    \item We present a self-training framework, STL that leverages structured rationales to jointly improve perception and reasoning, in contrast to prior approaches that emphasize reasoning alone.
    \item We assess our method across domains such as commonsense, language, and science, and present detailed ablations showing that STL boosts performance while avoiding the limitations of human or proprietary supervision.
\end{enumerate}
\section{Background and Related Work}
\textbf{Vision-Language Reasoning.} Reasoning \cite{zhang2023multimodal, cesista2024multimodal, xu2024llavacot} has been shown to play a critical role in enhancing the performance of Vision-Language Models (VLMs). While recent VLMs achieve strong results on general benchmarks \cite{liu2024improved, chen2024far}, effectively incorporating visual information into the reasoning process remains a persistent challenge, particularly for open-source models \cite{liu2023llava, liu2023improvedllava, bai2308qwen}. One direction has explored prompt-based strategies that assign functional roles to VLMs via system prompts, enabling modular step-by-step reasoning. For example, Cantor \cite{gao2024cantor} structures reasoning into context analysis followed by high-level feature generation, while CCoT \cite{mitra2024compositional} leverages scene graphs to capture object-attribute relations and guide a two-stage process of graph construction and reasoning.

Another direction relies on learning-based approaches that fine-tune VLMs on datasets containing multimodal reasoning chains \cite{xu2411llava, yao2024mulberry, thawakar2025llamav, shao2024visual}. Such datasets are typically curated using powerful teacher models (e.g. GPT-4o, Deepseek-R1) or through costly human annotations \cite{lu2022learn, chen2024far}, raising concerns about scalability. In contrast, our work avoids this reliance by enabling VLMs to enhance their reasoning abilities through self-learning, without requiring curated rationales.

\noindent\textbf{Self-Training Methods.} Self-training is a semi-supervised paradigm where a model improves by generating supervision from its own outputs. In Large Language Models (LLMs), it has been widely used to strengthen reasoning: models generate intermediate rationales (e.g. chain-of-thought) for unlabeled data, which are then reused as training signals in subsequent iterations. This iterative process enhances reasoning while reducing dependence on human annotations \cite{yuan2025selfrewardinglanguagemodels, chen2024self}. Seminal works have advanced LLM reasoning by sampling rationales, filtering them for correctness, and fine-tuning on positive samples \cite{zelikman2022star, hosseini2024v, yuan2023scaling}. By contrast, self-training for VLMs remains relatively underexplored.

In the video domain, Video-STaR \cite{zohar2024video} extends STaR \cite{zelikman2022star} by generating question-answer pairs from labeled video datasets for instruction tuning. More recently, R\textsuperscript{3}V \cite{cheng2025vision} explored self-training for reasoning in VLMs. However, unlike STL, R\textsuperscript{3}V relies on knowledge distillation from GPT-4o \cite{openai_gpt4o_2024} as a warm-up stage and requires additional bookkeeping to track responses that evolve from incorrect to correct across iterations.

Furthermore, both Video-STaR and R\textsuperscript{3}V omit explicit visual descriptions in their reasoning templates. STL differs by introducing a cost-effective reasoning template that integrates image descriptions, thereby improving perception and reasoning over successive iterations. Additionally, STL leverages discriminative learning to unlearn spurious correlations that otherwise lead to systematic errors.

\section{Method}

The Self-Taught Reasoner (STaR) algorithm \cite{zelikman2022star} is a seminal self-training approach to improve reasoning in LLMs. For relevance to our setting, we first describe its extension to Vision-Language Models (VLMs). We then introduce our proposed framework, \textbf{See-Think-Learn (STL)}, which addresses the limitations of STaR by jointly refining perception and reasoning. We assume access to a multiple-choice VQA dataset
$$
D = \{(I_i, x_i, C_i, a_i)\}_{i=1}^N,
$$
where $I_i$ is an image, $x_i$ a question, $C_i$ the candidate answers, and $a_i$ the correct answer. 

\subsection{STaR for Enhancing VLM Reasoning}
A simple adaptation of STaR to the VLM setting involves providing its self-training loop with multimodal input, allowing the model to perceive images and reason over text. A VLM $M$ maps the input $(I, x, C)$ to a rationale-answer pair:
$$
(r, \hat{a}) = M(I, x, C),
$$
where $r$ is a rationale in natural-language and $\hat{a}$ the predicted answer.
At iteration $n$, the model produces
$$
(r_i, \hat{a}_i) = M_{n-1}(I_i, x_i, C_i), \quad i = 1, \dots, N.
$$
We then partition these outputs into correct and incorrect predictions:
$$
D_n^+ = \{(I_i, x_i, C_i, r_i, a_i) \mid \hat{a}_i = a_i\},
$$
$$
D_n^- = \{(I_i, x_i, C_i, r_i, \hat{a}_i) \mid \hat{a}_i \neq a_i\}.
$$
In the STaR framework, both sets are used for retraining. Correct predictions ($D_n^+$) are directly fine-tuned, while incorrect predictions ($D_n^-$) undergo \emph{positive rationalization}: the model is given the gold answer $a_i$ and asked to produce a new rationale $\tilde{r}_i$ that supports it. These newly generated rationales are then filtered to retain only those that lead to the correct answer. This yields a corrected set:
$$
\tilde{D}_n^+ = \{(I_i, x_i, C_i, \tilde{r}_i, a_i) \mid (I_i, x_i, C_i, r_i, \hat{a}_i) \in D_n^-\}.
$$
The combined dataset
$$
D_n = D_n^+ \cup \tilde{D}_n^+
$$
is used to fine-tune the model $M$ to yield the model $M_{n}$.

By iteratively generating, correcting and retraining on rationales, STaR bootstraps reasoning ability without additional human supervision. However, when applied to VLMs, it faces two key limitations:
\begin{enumerate}
    \item \textbf{Perceptual grounding gap:} STaR’s generated rationales mix perception with reasoning, preventing the reasoning from being properly conditioned on perceptual inputs.
    \item \textbf{Noisy rationales from positive rationalization:} Because the gold answer is revealed, the model may generate superficially correct rationales while retaining flawed reasoning (Figure~\ref{fig:starvsours}).
\end{enumerate}

\begin{figure}
\centering
  \includegraphics [width=0.47\textwidth]{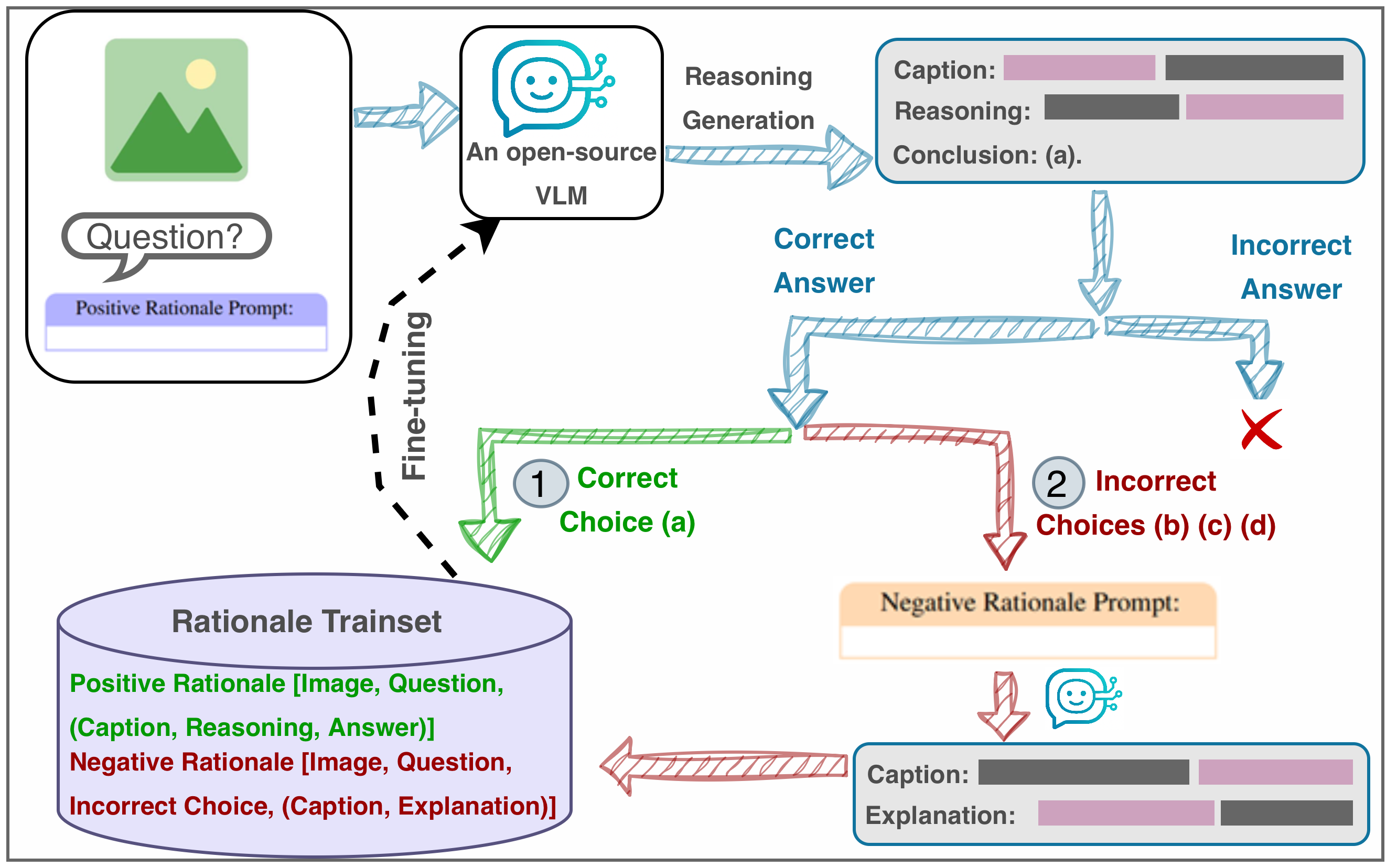}
\caption{\textbf{An overview of our detailed “See-Think-Learn” (STL) framework. } 
In this framework, each image-question pair with multiple choices, together with a positive rationale prompt, is fed into the VLM to generate a caption, reasoning, and conclusion. If the model predicts the correct answer, the tuple
[Image, Question, (Caption, Reasoning, Answer)]
is stored as a \textbf{Positive Rationale} in the Rationale Trainset. The remaining incorrect choices are used to generate negative rationalizations, producing a caption and an explanation of why the choice is incorrect, which are stored as \textbf{Negative Rationales} 
[Image, Question, Incorrect Choice, (Caption, Explanation)].
The VLM is then iteratively fine-tuned on this dynamically constructed Rationale Trainset.}

\label{model}
\end{figure}

\subsection{Self-Training Framework: See-Think-Learn (STL)}
To address these limitations, we propose \textbf{STL}, a self-training framework that integrates perception into the reasoning loop. STL differs from STaR in three key aspects:

\begin{figure*}[t]
\centering

\begin{minipage}{0.48\textwidth}
\begin{tcolorbox}[
  colback=white,
  colframe=blue!30,
  coltext=black,
  coltitle=black,
  boxrule=0.8pt,
  arc=5pt,
  height=5.7cm,  
  sharp corners=south,
  fontupper=\fontsize{7}{8}\selectfont\ttfamily,
  title=Positive Rationale Prompt:,
  before skip=8pt,
  after skip=8pt
]
You are an image based question-answering expert. \\
Given an image along with a multiple choice question, your task is to select the correct choice based on the image. \\
Your response should strictly follow the format with three specific sections: CAPTION, REASONING and CONCLUSION. Response:\\
\#\#\#CAPTION: [Provide a detailed description of the image, particularly emphasizing the aspects related to the question.]\\
\#\#\#REASONING: [Provide a detailed thought process to answer the question.]\\
\#\#\#CONCLUSION: [Provide the correct choice based on the reasoning.]\\
Question: \{question\_and\_choices\}\\
Response: 
\end{tcolorbox}
\end{minipage}
\hfill
\begin{minipage}{0.48\textwidth}
\begin{tcolorbox}[
  colback=white,
  colframe=orange!30,
  coltext=black,
  coltitle=black,
  boxrule=0.8pt,
  arc=5pt,
  height=5.7cm,  
  sharp corners=south,
  fontupper=\fontsize{7}{8}\selectfont\ttfamily,
  title=Negative Rationale Prompt:,
  before skip=8pt,
  after skip=8pt
]
You are an image based question-answering expert. \\
Given an image along with a multiple choice question and an answer, your task is to explain why the answer is wrong. \\
Your response should strictly follow the format with two specific sections: CAPTION and EXPLANATION. Response: \\
\#\#\#CAPTION: [Provide a detailed description of the image, particularly emphasizing the aspects related to the question.] \\
\#\#\#EXPLANATION: [Provide a detailed explanation for why the answer is wrong.] \\ 
Question: \{question\} \\
The correct choice is \{correct\_choice\}. \\
Explain why this answer is wrong: \{incorrect\_choice\} \\
Response: \\
\end{tcolorbox}
\end{minipage}

\caption{\textbf{Prompt templates} used for positive and negative rationalization in the STL framework.}
\label{prompts}
\end{figure*}
\vspace{0.5em}

\noindent\textbf{Structured rationale prompts:} We introduce a structured rationale prompt that follows a "see before thinking" approach. As shown in Figure~\ref{prompts}, our prompt has three parts: \textbf{(1) Caption}, which gives a detailed description of the image based on the question; \textbf{(2) Reasoning}, which involves a detailed thought process grounded in visual details; and \textbf{(3) Conclusion}, which gives the final answer based on reasoning. Specifically, the model is prompted to produce a tuple \( (d_i, r_i, \hat{a}_i) \), comprising an image description \( d_i \), a rationale \( r_i \), and a predicted answer \( \hat{a}_i \). 

\vspace{0.5em}
\noindent\textbf{Selective positive rationales:} Instead of correcting rationales for incorrect predictions, STL retains only high-quality rationales from correctly answered samples (obtained without revealing the gold answer), thereby reducing noise. 
Concretely, for each instance the model outputs
$$
(d_i, r_i, \hat{a}_i) = M(I_i, x_i, C_i),
$$
where $d_i$ is an image description, $r_i$ a rationale, and $\hat{a}_i$ the predicted answer. We retain only the correct predictions:
$$
D^{\text{pos}}_n = \{(I_i, x_i, C_i, d_i, \hat{r}_i^+, a_i) \mid \hat{a}_i = a_i\}.
$$
This dataset is used for iterative fine-tuning, encouraging the model to improve both perception and reasoning.

\vspace{0.5em}

\noindent\textbf{Discriminative negative rationales:} Rationales from correct samples capture only one side of the reasoning spectrum. To emulate the human strategy of reflective learning, where incorrect options are critically analyzed, we further augment the training data with \textit{negative rationales}: explanations for why incorrect choices are wrong. This enhances the model's ability to distinguish the correct answer from the alternatives. We now describe this process in detail in the following subsection.

\subsection{Negative Rationalization: Discriminative Learning}
To further strengthen reasoning, STL introduces \textbf{negative rationalization}. For this, we focus only on samples that the model previously answered correctly. We assume that a correct prediction means that the model has a good understanding of the image and the question. Using this confidence, we prompt the model (see Figure~\ref{prompts}) to first describe the image ($d_i$), then explain why each incorrect option $c \in C_i \setminus {a_i}$ is not valid. These explanations are generated with the correct answer $a_i$ included in the prompt to guide accurate reasoning. Each explanation is stored as a tuple $(I_i, x_i, C_i, c, d_i, \hat{r}_{i,c}^{-})$, where $\hat{r}_{i,c}^{-}$ is the negative rationale for option $c$.

Later, when used for training, the gold answer is withheld from the prompt, forcing the model to reason independently about distractors. This yields the negative rationale dataset:
\begin{equation}
\begin{aligned}
D^{\text{neg}}_n = \{(I_i, x_i, C_i, c, d_i, \hat{r}_{i,c}^-) \; \mid \; 
&(I_i, x_i, C_i, d_i, \hat{r}_i^+, a_i) \in \\ &D^{\text{pos}}_n,c \in C_i \setminus \{a_i\} \}.
\end{aligned}
\end{equation}

Fine-tuning on the combined dataset $D^{\text{pos}}_n \cup D^{\text{neg}}_n$
enhances three complementary abilities:
(1) perceptual grounding via structured descriptions, (2) reasoning ability via positive rationales and (3) discriminative capability via negative rationales, allowing the model not only to generate coherent rationales for correct answers, but also to critically assess and reject misleading or implausible alternatives. Algorithm~\ref{alg:stl} summarizes the entire STL procedure and Figure~\ref{model} illustrates its workflow.
\begin{algorithm}
\caption{\textsc{STL: See–Think–Learn}}
\label{alg:stl}
\begin{footnotesize}
\begin{algorithmic}[1]
\Require Pretrained VLM $M$; MCQ dataset $\mathcal{D} = \{(I_i, x_i, \mathcal{C}_i, a_i)\}_{i=1}^D$
\State $M_0 \gets M$; \quad $n \gets 0$
\Repeat 
  \State $n \gets n + 1$
  
  \State \textbf{(A) Inference: Generate positive rationales (description, reasoning and prediction)}
  \ForAll{$(I_i, x_i, \mathcal{C}_i, a_i) \in \mathcal{D}$}
    \State $(d_i, \hat r^+_i, \hat a_i) \gets M_{n-1}[\texttt{posprompt}, I_i, x_i, \mathcal{C}_i]$
  \EndFor
  
  \State \textbf{(B) Construct positive rationale dataset from correct predictions}
  \State $\mathcal{D}^\text{pos}_n \gets \{(I_i, x_i, C_i, d_i, \hat r^+_i, a_i) \mid \hat a_i = a_i\}$
  
  \State \textbf{(C) Inference: Generate negative rationales (description and explanation)}
  \ForAll{$(I_i, x_i, C_i, d_i, \hat r^+_i, a_i) \in \mathcal{D}^\text{pos}_n$}
    \ForAll{$c \in \mathcal{C}_i \setminus \{a_i\}$}
      \State $(d_i, \hat r^-_{i,c}) \gets M_{n-1}[\texttt{negprompt}, I_i, x_i, C_i, a_i, c]$
    \EndFor
  \EndFor
  
  \State \textbf{(D) Construct negative rationale dataset}
  \State $\mathcal{D}^\text{neg}_n \gets \{(I_i, x_i, C_i, c, d_i, \hat r^-_{i,c}) \mid c \in \mathcal{C}_i \setminus \{a_i\} \}$

  \State \textbf{(E) Combine and fine-tune}
  \State $\mathcal{D}_n \gets \mathcal{D}^\text{pos}_n \cup \mathcal{D}^\text{neg}_n$
  \State $M_n \gets \texttt{train}(M, \mathcal{D}_n)$
  
\Until{converged}
\end{algorithmic}
\end{footnotesize}
\end{algorithm}

\section{Experiments}
To assess the effectiveness of our method, we conducted comprehensive evaluations across four knowledge domains using the LLaVA-v1.5-7B \cite{liu2023llava} model. To examine the generalizability of our method across different VLMs, we further conducted experiments with the Qwen2.5-VL-7B-Instruct model \cite{qwen2.5-VL}. We start by detailing the datasets and baseline methods used for comparison, implementation details followed by the quantitative results and qualitative analysis. 

\subsection{Datasets} 
We evaluated our method across four domains, namely commonsense, natural science, language science, and social science, using samples drawn from the M3CoT dataset \cite{chen-etal-2024-m3cot}, which provides multi-domain, multiple-choice visual question-answering tuples paired with human-annotated rationales. The availability of these rationales allows a direct comparison between our approach and human-annotated reasoning. The commonsense domain assesses reasoning about physical, social, and temporal aspects depicted in images; natural science focuses on visually grounded questions in physics, chemistry, and biology; social science addresses topics related to geography, economics, and cognitive science; and language science encompasses questions involving figurative language, grammar, and reading comprehension.

\begin{table}
  \centering
  \small
  \renewcommand{\arraystretch}{1.25}
  \setlength{\tabcolsep}{2pt}
  \rowcolors{2}{gray!10}{white}
\caption{\textbf{Performance Comparison on M3CoT Evaluation Splits on LLaVA \cite{liu2023llava}.} Accuracy of various baselines along with the proposed STL across the four domains.}

  \label{tab:main_results}
  \begin{tabular}{p{1.4cm}ccccc}
    \toprule
    \rowcolor{gray!20} 
    \textbf{Method} & \textbf{\begin{tabular}[c]{@{}c@{}}Common\\ sense\end{tabular}} & \textbf{\begin{tabular}[c]{@{}c@{}}Natural-\\ Science\end{tabular}} & \textbf{\begin{tabular}[c]{@{}c@{}}Language-\\ ‑Science\end{tabular}} & \textbf{\begin{tabular}[c]{@{}c@{}}Social‑\\ ‑Science\end{tabular}} & \textbf{Average} \\ 
    \midrule
    \multicolumn{6}{c}{\textbf{Zero-Shot Methods}} \\
    Direct VQA & 57.58 & 36.40 & 45.02 & 29.62 & 42.16 \\
    CoT & 54.94 & 35.5 & 35.54 & 24.68 & 37.66 \\
    Positive Prompt (Fig. \ref{prompts}) & 53.40 & 33.20 & 40.28 & 27.55 & 38.61 \\
    \midrule
    \multicolumn{6}{c}{\textbf{Direct SFT}} \\
    Direct SFT & 60.22 & 46.10 & 46.92 & 34.24 & 46.87 \\
    \midrule
    \multicolumn{6}{c}{\textbf{Self-Training Methods}} \\
    STaR\cite{zelikman2022star} & 64.98 & \textbf{53.90} & 48.82 & 41.88 & 51.21 \\
    R\textsuperscript{3}V\cite{cheng2025vision} & 62.64 & - & 45.97 & - & - \\
    Ours & \textbf{67.19} & 50.45 & \textbf{55.92} & \textbf{43.79} & \textbf{54.34} \\
    \bottomrule
  \end{tabular}
\end{table}
\subsection{Baselines}
We compare our approach against several strong baselines. For zero-shot evaluation, we assess the base model’s performance under different prompting strategies, including direct answer prompting, Chain-of-Thought (CoT) prompting, and our structured rationale prompt. We also include a direct SFT baseline, where the model is fine-tuned on (image, question, answer) tuples using direct prompting, instructing it to predict the answer without generating an intermediate rationale. Finally, we compare our method with two state-of-the-art self-training approaches: STaR~\cite{zelikman2022star} and R\textsuperscript{3}V~\cite{cheng2025vision}.

\subsection{Implementation Details}
We fine-tuned both the LLaVA-v1.5-7B~\cite{liu2023llava} and Qwen2.5-VL-7B-Instruct~\cite{qwen2.5-VL} models using LoRA with rank \( r = 128 \) and scaling factor \( \alpha = 256 \). To support memory-efficient training, we used DeepSpeed ZeRO-3 and enabled gradient checkpointing. Training was performed for one epoch with a batch size of 8 and 3 gradient accumulation steps under a cosine learning-rate schedule. All experiments were run on a single NVIDIA A6000 GPU (48~GB VRAM). The self-training methods were executed for 6–7 iterations, and we report test-set accuracy for all results.

\begin{table}
  \centering
  \small
  \renewcommand{\arraystretch}{1.25}
  \setlength{\tabcolsep}{2pt}
  \rowcolors{2}{gray!10}{white}
\caption{\textbf{Performance Comparison on M3CoT Evaluation Splits for Qwen \cite{qwen2.5-VL}.} Accuracy of various baselines along with the proposed STL across two domains.}

  \label{tab:main_results_qwen}
  \resizebox{\columnwidth}{!}{
  \begin{tabular}{p{3.5cm}ccc}
    \toprule
    \rowcolor{gray!20} 
    \textbf{Method} & \textbf{Commonsense}  & \textbf{Language Science} &  \textbf{Average} \\ 
    \midrule
    \multicolumn{4}{c}{\textbf{Zero-Shot Methods}} \\
    Direct VQA & 82.20  & 72.51 & 77.36 \\
    CoT & 80.00 & 57.34 & 68.67  \\
    Positive Prompt (Fig. \ref{prompts}) & 80.48 & 73.46 & 76.97 \\
    \midrule
    \multicolumn{4}{c}{\textbf{Direct SFT}} \\
    Direct SFT  & 82.42 & 79.15 &  80.79 \\
    \midrule
    \multicolumn{4}{c}{\textbf{Self-Training Methods}} \\
    STaR\cite{zelikman2022star} & 81.54 &  80.57 & 81.06  \\
    R\textsuperscript{3}V\cite{cheng2025vision} & 80.44 & 74.88 & 77.66 \\
    Ours & \textbf{84.32} & \textbf{86.41} & \textbf{85.36} \\
    \bottomrule
  \end{tabular}
  }
\end{table}

\begin{figure*}
\definecolor{DarkGreen}{RGB}{1,100,3}
\definecolor{DarkRed}{RGB}{100,20,0}
\centering
  \includegraphics [width=\textwidth]{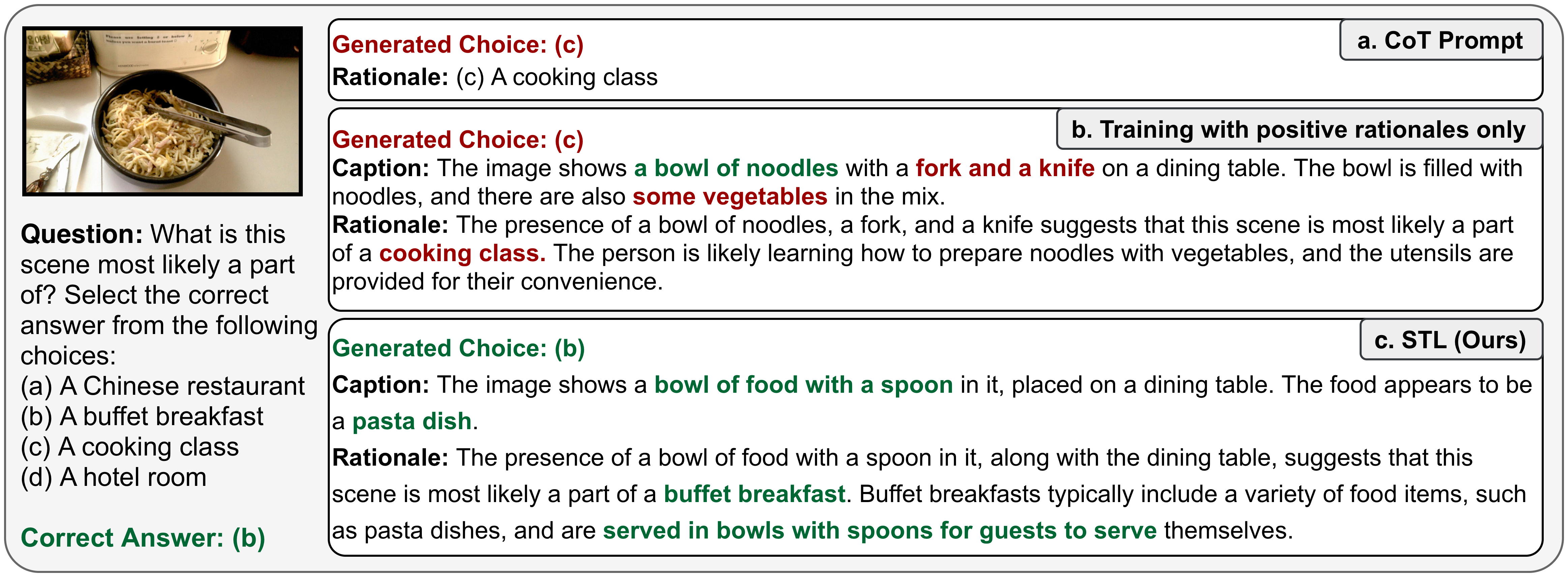}
\caption{\textbf{Comparison of our ``See-Think-Learn” (STL) framework with CoT Prompting.} The example is taken from the Commonsense Split of M3CoT Dataset \cite{chen-etal-2024-m3cot}. Unlike CoT prompting a., our STL framework (b. and c.) effectively generates a detailed description and accurate reasoning for the image by leveraging the proposed Positive and Negative Rationale Prompts. In a., the answer is incorrect, and the image description is missing. In b., although a detailed description is provided, it is inaccurate. For example, it mentions a \textcolor{red}{“fork”} and \textcolor{red}{“knife”} that are not present in the image. In contrast, c. produces both the correct answer and an accurate description, capturing key elements such as \textcolor{DarkGreen}{“serve”} and \textcolor{DarkGreen}{“buffet”}. Q: Question; O: Options; }
\label{model_analysis}
\end{figure*}
\section{Quantitative Evaluation}
We evaluated STL against several baselines using accuracy, with results on LLaVA presented in Table \ref{tab:main_results}. Zero-shot prompting struggles with multimodal reasoning, and Chain-of-Thought (CoT) prompting performs worse than direct prompting (Direct VQA). The higher accuracy observed in the Direct VQA setting compared to CoT prompting indicates the inherent shortcut-seeking behavior of MLLMs. When prompted to reason before answering, models such as LLaVA, which possess limited reasoning capabilities, often generate incorrect rationales, ultimately leading to incorrect answers. Direct SFT, which trains the model to answer questions without generating rationales, achieves a modest 4\% gain over zero-shot methods. 

Self-training methods leverage the base model’s weak reasoning to bootstrap reasoning datasets, achieving substantial gains. STL performs best, surpassing STaR by 3\% and R\textsuperscript{3}V by 7\%, producing higher-quality rationales than the baselines. Notably, after STL training, the model can generate coherent rationales for new problems, contributing to its improved performance.

Interestingly, STaR outperforms STL on the Natural Science dataset. This is largely due to STaR’s sample generation strategy: besides using high-quality positive samples, it employs positive rationalization by reprompting incorrect answers with hints emphasizing the correct choice. This enables the model to reach correct answers without true reasoning, a shortcut that can produce inconsistent or misaligned explanations. In contrast, STL relies solely on correctly answered samples and applies negative rationalization for augmentation, encouraging the model to differentiate correct from incorrect reasoning. Although STaR benefits from a greater exposure to training data, STL prioritizes robust, genuine reasoning. Figure \ref{fig:starvsours} illustrates an example of shortcut learning in STaR.

\begin{table}[t]
  \centering
  \scriptsize
  \renewcommand{\arraystretch}{1.2}
  \setlength{\tabcolsep}{4pt}
  \caption{\textbf{Comparison with human-annotated rationale training.}\\
  Evaluation of STL vs. M3CoT \cite{chen-etal-2024-m3cot} rationales on Language-Science and Commonsense splits.}
  \label{tab:m3cot_rationale_comparison}
  \resizebox{\columnwidth}{!}{%
  \begin{tabular}{lccc}
    \toprule
    \rowcolor{gray!20}
    \textbf{Model} & \textbf{Language Science} & \textbf{Commonsense} & \textbf{Average} \\ 
    \midrule
    LLaVA (Human) & \textbf{66.82} & 60.44 & \textbf{63.63} \\
    LLaVA (Ours) & 55.92 & \textbf{67.19} & 61.56 \\
    \midrule
    Qwen (Human) & 72.98 & 78.68 & 75.83 \\
    Qwen (Ours) & \textbf{86.41} & \textbf{84.32} & \textbf{85.36} \\
    \bottomrule
  \end{tabular}}
\end{table}

\begin{table}
  \centering
  \small
  \renewcommand{\arraystretch}{1.25}
  \setlength{\tabcolsep}{8pt}
  \rowcolors{2}{gray!10}{white}
  \caption{\textbf{Ablation Study.} W/O: Without, W/: With, Neg: Negative Rationalization, Cap: Structured Rationale Prompt}
  \label{tab:ablation}
  \resizebox{\columnwidth}{!}{
  \begin{tabular}{lcc}
    \toprule
    \rowcolor{gray!20}
    \textbf{Method} & \textbf{Language Science} & \textbf{Commonsense} \\
    \midrule
    W/O (Cap+Neg)      & 46.44 & 59.56 \\
    W/O Neg   & 48.34 & 64.62 \\
    Ours (W/ (Cap+Neg))      & \textbf{55.92} & \textbf{67.19} \\
    \bottomrule
  \end{tabular}
  }
\end{table}

To further assess the generalizability of our approach to stronger, modern VLMs, we applied the STL framework to Qwen2.5-VL-7B~\cite{qwen2.5-VL} on the Commonsense and Language Science domains. Compared to LLaVA, the Qwen model exhibits superior reasoning abilities, as reflected in Direct VQA performance. However, shortcut learning is still evident: performance drops noticeably when the model is prompted to generate reasoning before answering.
Fine-tuning directly on the answers improves overall performance by 3\%. As expected, self-training methods provide much larger gains, demonstrating that STL can also be used to enhance the reasoning capabilities of stronger models.

\begin{figure*}
  \centering
  \includegraphics[width=\textwidth]{figures/starvsours.png}
  \caption{\textbf{Qualitative Comparison on Natural Science Domain.} Qualitative analysis shows that STL (ours) produces more coherent and logically consistent explanations than STaR, indicating deeper understanding and more faithful reasoning.} 
  \label{fig:starvsours}
\end{figure*}

\noindent\textbf{Comparison with human-annotated rationales:}
To compare the reasoning quality, we use the human-annotated rationales from M3CoT. Table \ref{tab:m3cot_rationale_comparison} compares models fine-tuned on these human-annotated rationales with the models fine-tuned using See-Think-Learn (STL) self-generated rationales, across the Language Science and Commonsense domains. Although human-annotated rationales remain higher in quality, STL-generated samples achieve competitive or superior performance, demonstrating that STL can approximate human-level reasoning while providing a scalable alternative to manual annotation.

\begin{figure}
  \centering
  \includegraphics[width=.47\textwidth]{figures/sa_wacv_oneline.png}
  \caption{\textbf{Comparison of Preferred Rationale Counts across Domains.} Each subplot displays the number of times the rationale generated by each method (STaR and STL) was preferred (out of 150) within a domain. Across all domains, the reasoning generated by STL is preferred for more samples than that of STaR, highlighting its superior quality.}
  \label{fig:bar}
\end{figure}

\subsection{Ablation Studies}
We perform two ablation studies to assess the impact of each component of the proposed STL framework on the language science and commonsense domains. The W/O (Cap+Neg) condition represents our STL framework implemented without the proposed structured rationale prompt and negative rationalization i.e. this setup utilizes a prompt without caption and fine-tunes on just the correctly answered rationales.  The W/O Neg condition excludes only the negative rationalization component from our framework. As shown in Table~\ref{tab:ablation}, removing both the structured rationale prompt and negative rationalization results in noticeably lower performance, mainly due to the loss of enhanced perception and discriminative learning. Introducing the structured rationale prompt significantly improves accuracy by guiding the model to carefully analyze the image before reasoning, thereby enabling iterative refinement of its perception during training. Furthermore, incorporating negative rationalization augmentation yields an additional performance gain, as the model learns to better discriminate between correct and incorrect answers.

In Figure \ref{model_analysis}, we show a qualitative example, demonstrating the effect of the positive rationale template and discriminative learning through negative rationale template. The image depicts a bowl-based serving setup and the correct answer is \emph{buffet breakfast}. Figure \ref{model_analysis} a. shows the output of the model with CoT prompt (No description). The model fails to generate any rationale and answers the question incorrectly. When fine-tuning the model using self-training with Positive rationales (STL template) only, the model is able to generate a description which is partially correct leading to the same incorrect prediction. By contrast, STL fine-tuned using both positive and negative rationales corrects the answer and grounds reasoning in verifiable cues, such as a \textit{bowl of food with a spoon} and self-serve presentation, while avoiding nonexistent items. This suggests that STL improvements stem not from longer rationales but from the complementary effect of negative rationales, which suppress spurious tokens and guide the model toward subtle, task-relevant visual cues, enhancing both accuracy and rationale fidelity.

\subsection{Qualitative Results} 
\textbf{Qualitative Example on Commonsense:}
Figure \ref{fig:qual} shows an example from the commonsense domain with reasoning generated by STaR, R\textsuperscript{3}V, and STL. Both STaR and R\textsuperscript{3}V tend to overlook perceptual details and produce short explanations that resemble descriptive answers rather than grounded reasoning. In contrast, STL attends to contextual cues such as the plants in the background and the dirt area, enabling it to generate a more comprehensive rationale that leads to the correct answer.\\
\textbf{Qualitative Example of Natural Science Domain}:
Figure \ref{fig:starvsours} provides a qualitative comparison of reasoning outputs on the Natural Science domain, highlighting differences between STaR and STL. In the example, the model is asked to identify which animal is adapted to be camouflaged in a sandy desert.
The base model out of the box when prompted using CoT prompt is able to generate good reasoning but still end up predicting the incorrect choice demonstrating the inherent biases the model has when predicting the answers. STaR (with Positive Rationalization) encourages to learn incorrect reasoning. It mistakenly highlights the polar bear as camouflaged, and its rationales are inconsistent with the final answer, reflecting limited understanding of the visual context. In contrast, STL produces structured and logically consistent rationales. Using positive and negative rationalization, STL correctly identifies the lizard as adapted for camouflage and provides clear, step-by-step explanations. The caption accurately describes the scene, the reasoning links the observation to the correct choice, and the conclusion aligns perfectly with the rationale. The negative rationalization additionally reinforces the distinction by explaining why the polar bear is not the correct answer.

Overall, this example illustrates that STL generates more faithful and coherent explanations compared to STaR, indicating deeper understanding and stronger reasoning capabilities. Such qualitative improvements complement the quantitative gains observed in accuracy and suggest that STL promotes robust reasoning rather than relying on shortcuts. This also suggests that while STaR may achieve marginally better accuracy in some domains due to shortcut learning, STL fosters deeper understanding and more faithful reasoning. More qualitative samples are provided in the Supplementary Material.

\section{Subjective Analysis}
To assess rationale quality, we conducted a subjective evaluation comparing reasoning generated by STL and STaR for the LLaVA model. We randomly sampled 150 rationales per domain from questions both methods answered correctly and asked three annotators to compare them. Each annotator was shown the image, the corresponding question, and reasoning from both methods, without knowing which model produced them. Annotators ranked the two reasoning, and to ensure consistency, all assessed the same set of samples within each domain. Further details of the annotation procedure are provided in the Supplementary Material.

On average, STL reasoning were preferred 35\% more often than those from STaR. Figure \ref{fig:bar} summarizes these results, showing that across all domains, STL was consistently chosen more frequently. These findings demonstrate STL’s effectiveness in producing rationales that better align with human preferences and provide higher-quality reasoning compared to existing methods.

\section{Conclusion}
We present the See–Think–Learn (STL) framework, a self-training strategy that enables vision–language models to enhance multimodal reasoning without costly human-annotated rationales. By leveraging structured prompts with both positive and negative rationales, STL improves visual understanding and discriminative learning. The framework achieves strong empirical performance across tasks, demonstrating that models can effectively learn from their own generated perceptions and explanations, laying the groundwork for more advanced multimodal reasoning.

\small \bibliographystyle{ieeenat_fullname} \bibliography{main}

\newpage
\appendix

\vspace*{2em}

\begin{center}
    {\LARGE \bfseries See, Think, Learn: A Self-Taught Multimodal Reasoner}\\[2em]

    {\large \bfseries Supplementary Materials}\\[1em]

\end{center}

\section{Visualization}
\begin{figure*}
  \centering
  \includegraphics[width=.72\textwidth]{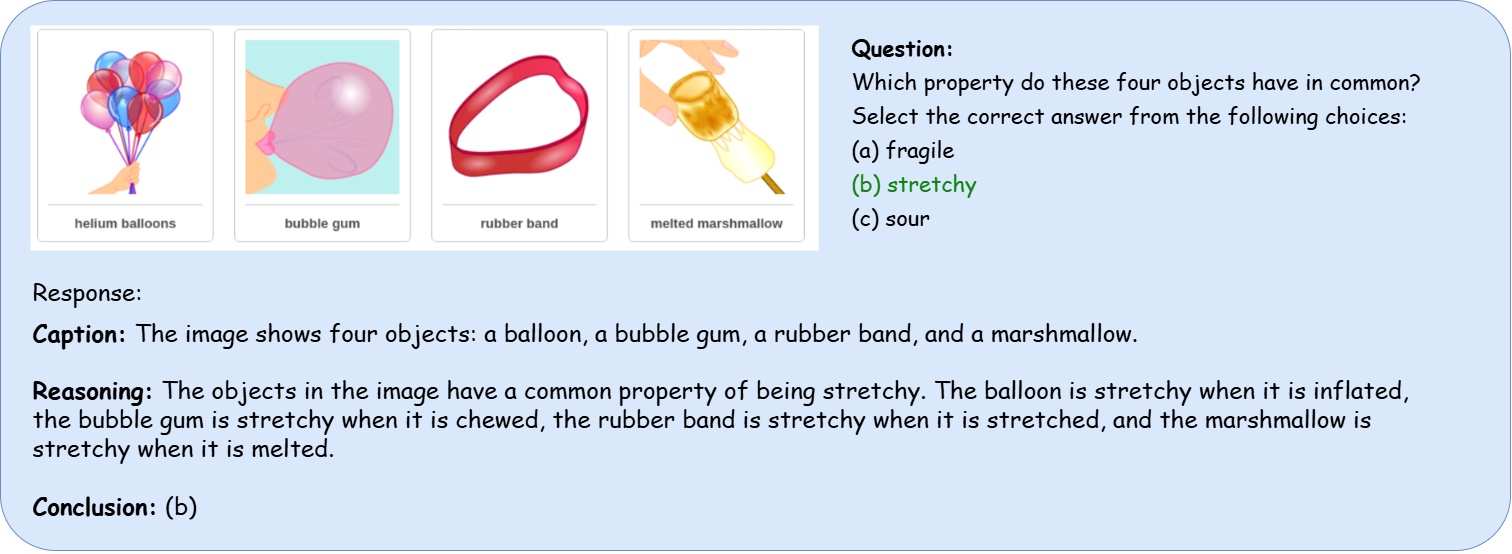}
  \includegraphics[width=.72\textwidth]{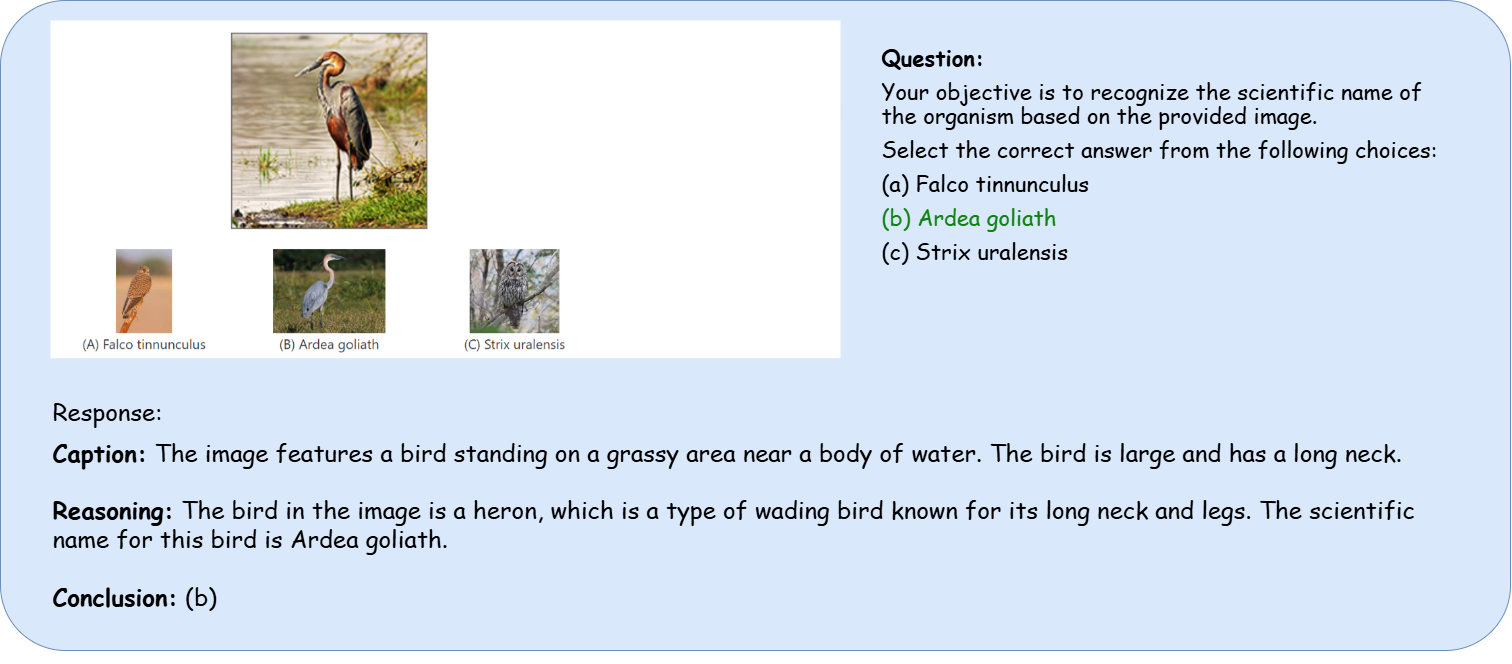}
  \includegraphics[width=.72\textwidth]{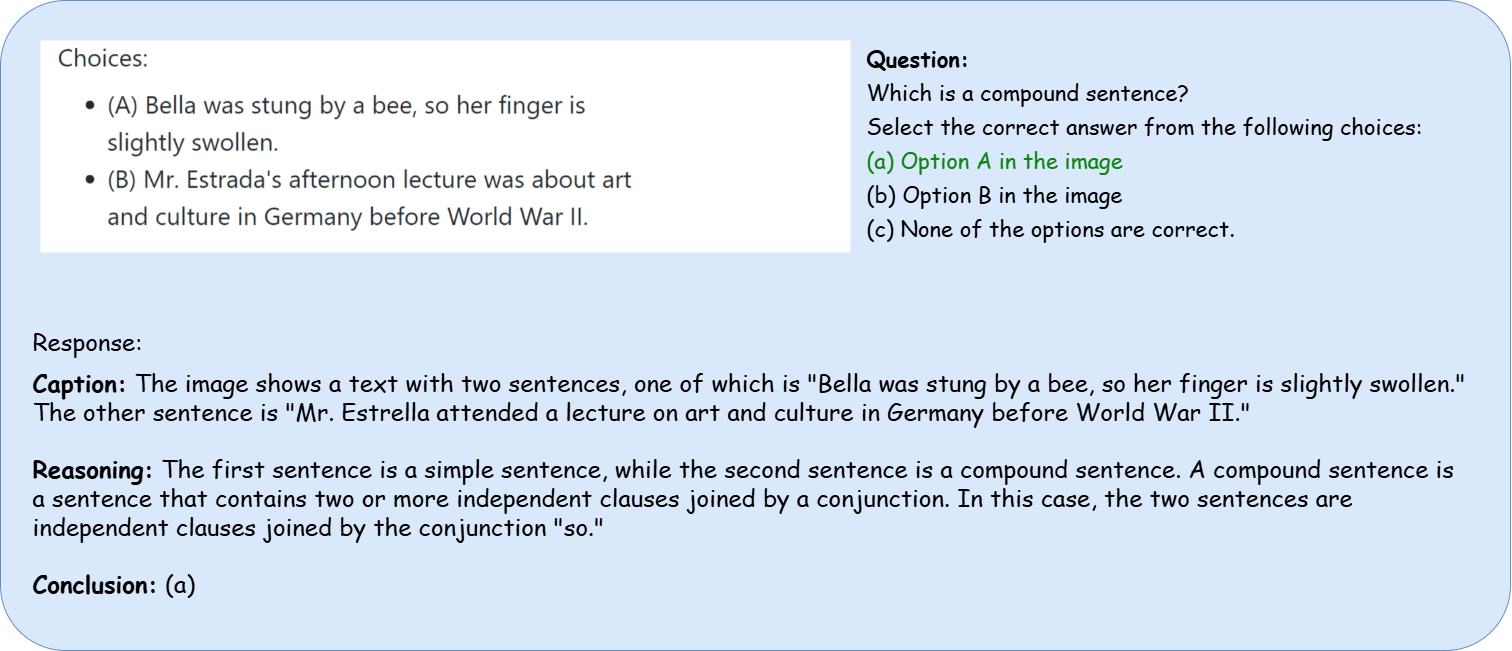}
\includegraphics[width=.72\textwidth]{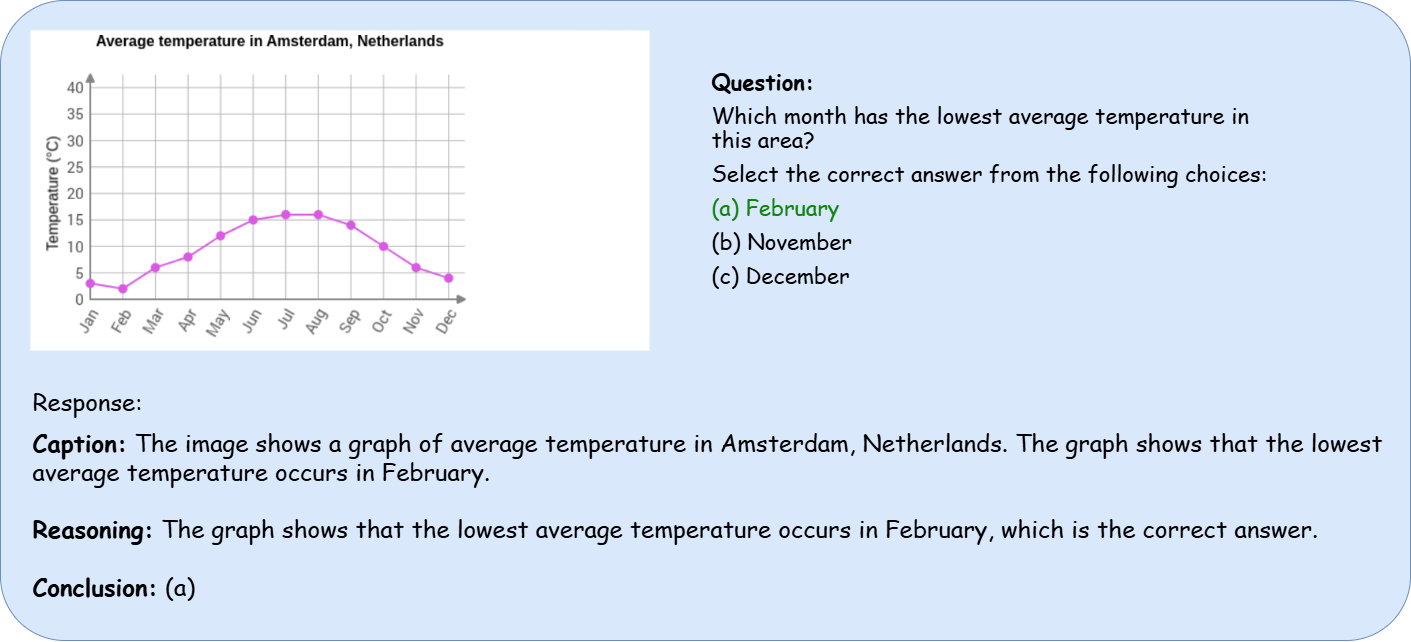}
  \caption{\textbf{Response for Commonsense, Natural Science, Language Science, Social Science Domains. (top to bottom)}. The response of our STL framework is a detailed caption and reasoning along with the answer. } 
  \label{fig:co}
\end{figure*}

\begin{figure*}
    \centering
    \begin{subfigure}[b]{0.58\textwidth}
        \includegraphics[width=\textwidth]{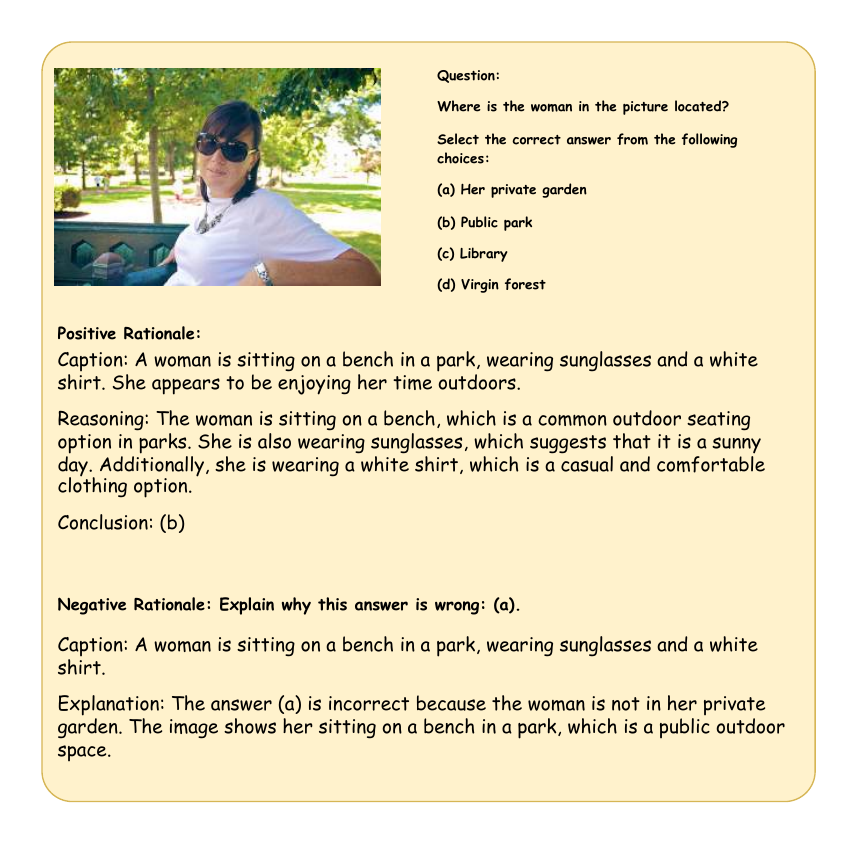}
    \end{subfigure}
    \hfill
    \begin{subfigure}[b]{0.58\textwidth}
        \includegraphics[width=\textwidth]{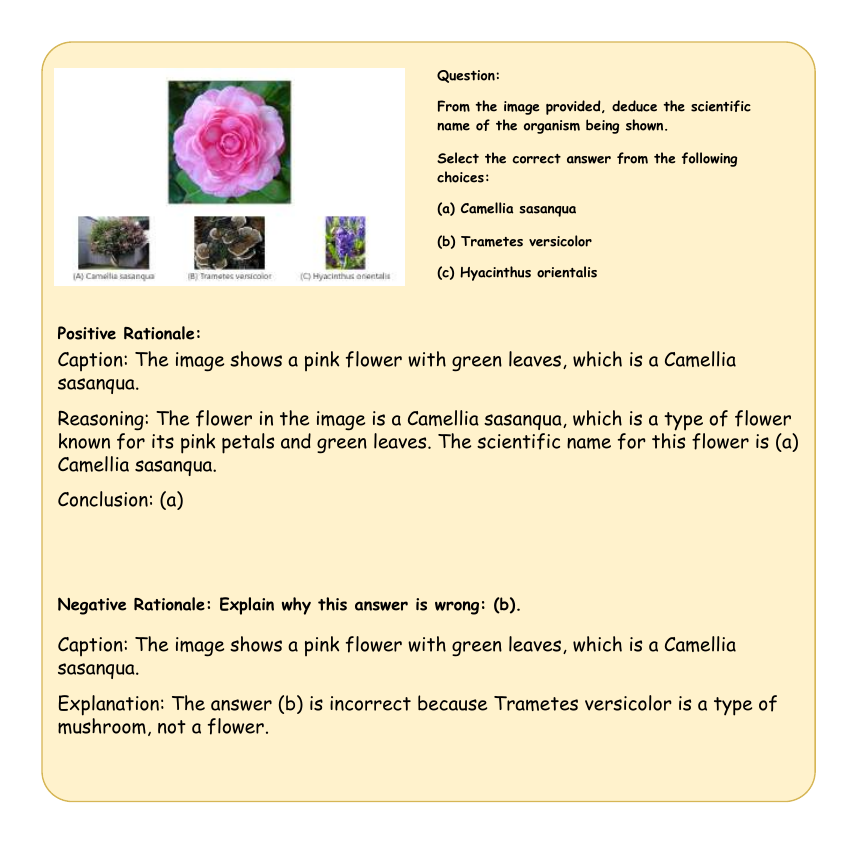}
    \end{subfigure}
    \vspace{2em}
    \caption{\textbf{Visualization of Positive and Negative Rationales from the proposed STL.} The examples illustrate correct identification and reasoning for the chosen answer and rejection of an incorrect alternative.}
    \label{fig:qualt_full}
\end{figure*}

Figure \ref{fig:co} illustrate the test set responses of domain-specific models trained using the proposed STL approach. As shown, the quality of reasoning significantly improves when preceded by high-quality image descriptions. During fine-tuning, the model is explicitly guided to enhance its perceptual understanding first. Subsequently, our structured rationale prompt directs the model to reason based on this refined perception. This sequential guidance, first enhancing perception and then structuring reasoning, leads to notable improvements in both components, ultimately resulting in more accurate answers. Furthermore, Figure \ref{fig:qualt_full} illustrates the positive and negative rationales produced by the model trained using the proposed STL framework.

\section{Additional Implementation Details}
\subsection{Structured Rationale Prompt: See-Before-Thinking}

Visual question answering requires both understanding what’s in the image and reasoning about it. The model needs to correctly interpret visual details and then make higher-level inferences to produce meaningful answers. However, standard chain-of-thought prompting methods, like the commonly used "Let's think step by step", often let the model skip proper visual grounding. In these cases, the model may rely on statistical patterns or shortcuts instead of focusing on the actual image, which can lead to plausible but incorrect answers.

To solve this problem, we introduce a structured rationale prompt that follows a "see before thinking" approach. As shown in Figure~\ref{prompts}, our method has three steps: \textbf{(1) Caption}, which gives a detailed description of the image based on the question; \textbf{(2) Reasoning}, which involves a step-by-step thought process grounded in visual details; and \textbf{(3) Conclusion}, which gives the final answer based on the reasoning.

This step-by-step structure encourages the model to look at the image first before starting to reason. By anchoring the reasoning in visual evidence, the model is more likely to give accurate and relevant answers, reducing its tendency to rely on unrelated patterns and improving performance on vision-based tasks.
\begin{figure}
  \centering
  \includegraphics[width=\columnwidth,height=0.14\textheight]{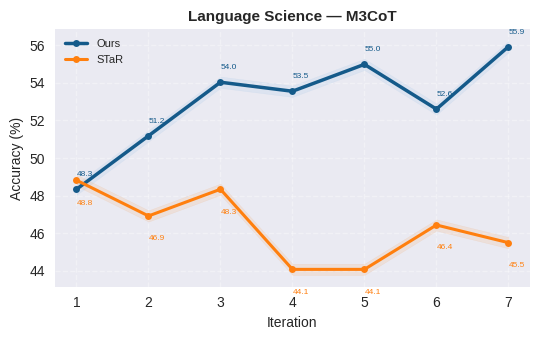}\par
  \includegraphics[width=\columnwidth,height=0.14\textheight]{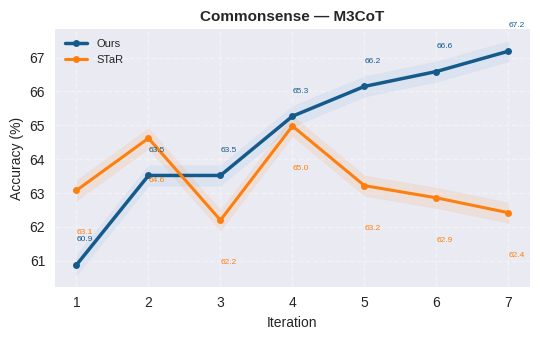}\par
  \includegraphics[width=\columnwidth,height=0.14\textheight]{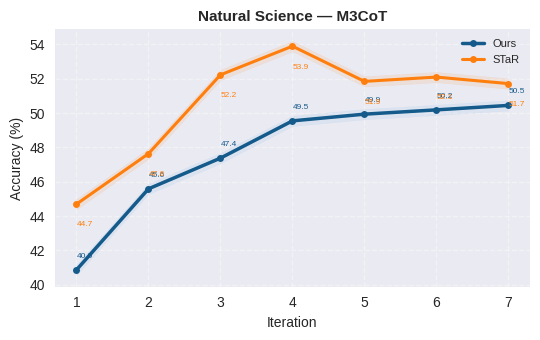}\par
  \includegraphics[width=\columnwidth,height=0.14\textheight]{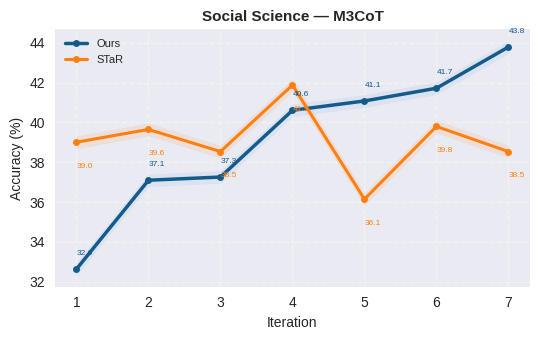}
  \caption{Comparison of test accuracy of STL (Ours) with STaR over 7 iterations.}
  \label{iteration_comparisons}
\end{figure}

\subsection{Training Setup}
In the original STaR framework, few-shot prompting is used to reduce noise in the generated bootstrapped rationale set. Likewise, R3V mitigates such noise by building the implementation and evaluation pipeline on a GPT-distilled baseline, where GPT-4o supplies high-quality CoT rationales for a small subset of each dataset. In contrast, we do not use few-shot prompting for STaR in our comparisons, nor do we adopt any GPT-based distillation as in R3V. This choice stems from the limited few-shot prompting capabilities of LVLMs and our core principle that \textit{quality outweighs quantity}: careful selection of high-quality rationales is more effective than relying on expensive GPT-generated warm-up annotations for each domain.

\subsection{Training Dynamics}
Figure \ref{iteration_comparisons} compares the trend of test accuracy of our STL framework with STaR over 7 iterations on the four domain splits of M3CoT dataset.

\begin{figure*}
    \centering
    \begin{subfigure}[b]{0.49\textwidth}
        \includegraphics[width=\textwidth]{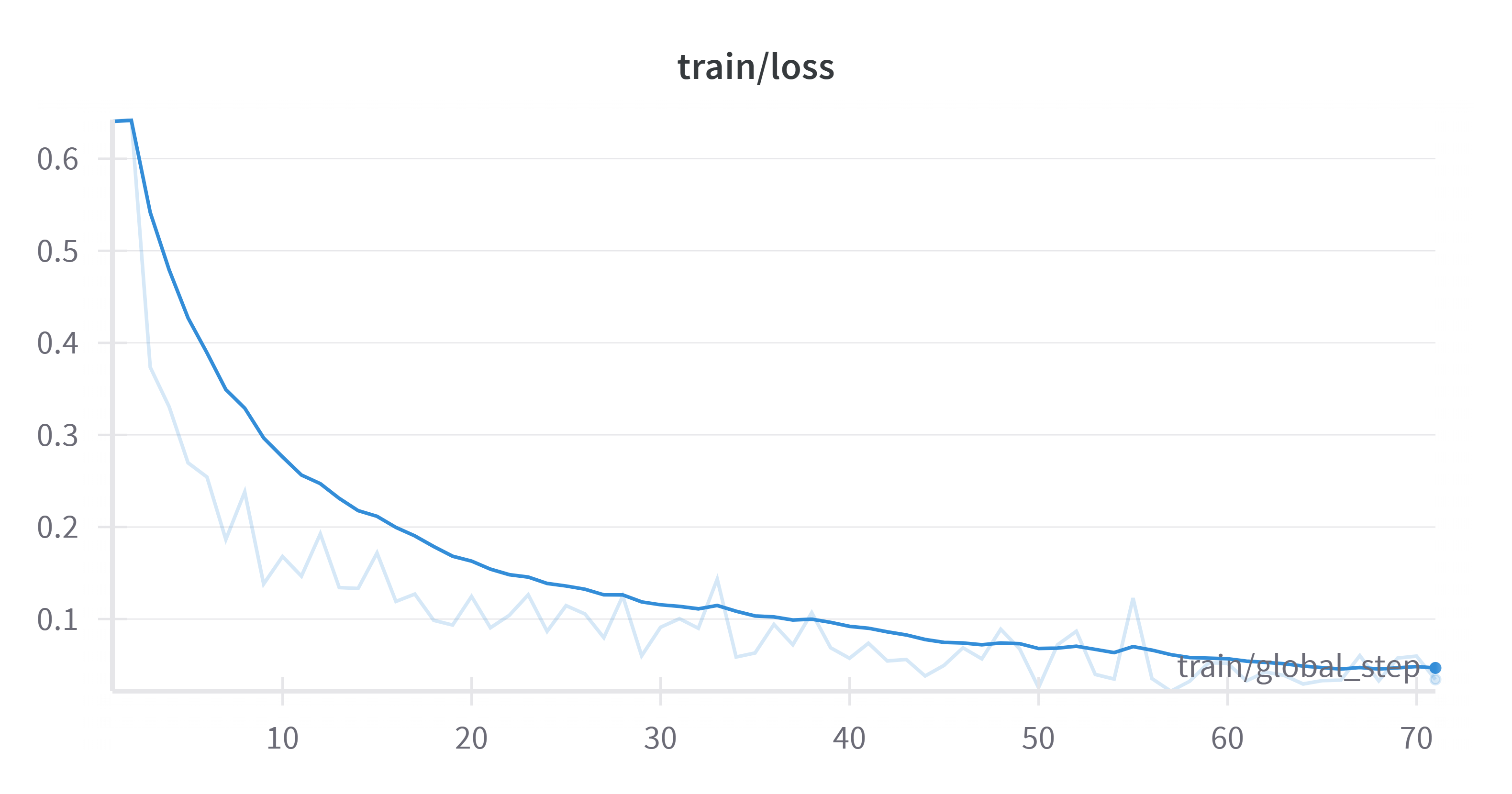}
        \caption{Training Loss}
    \end{subfigure}
    \hfill
    \begin{subfigure}[b]{0.49\textwidth}
        \includegraphics[width=\textwidth]{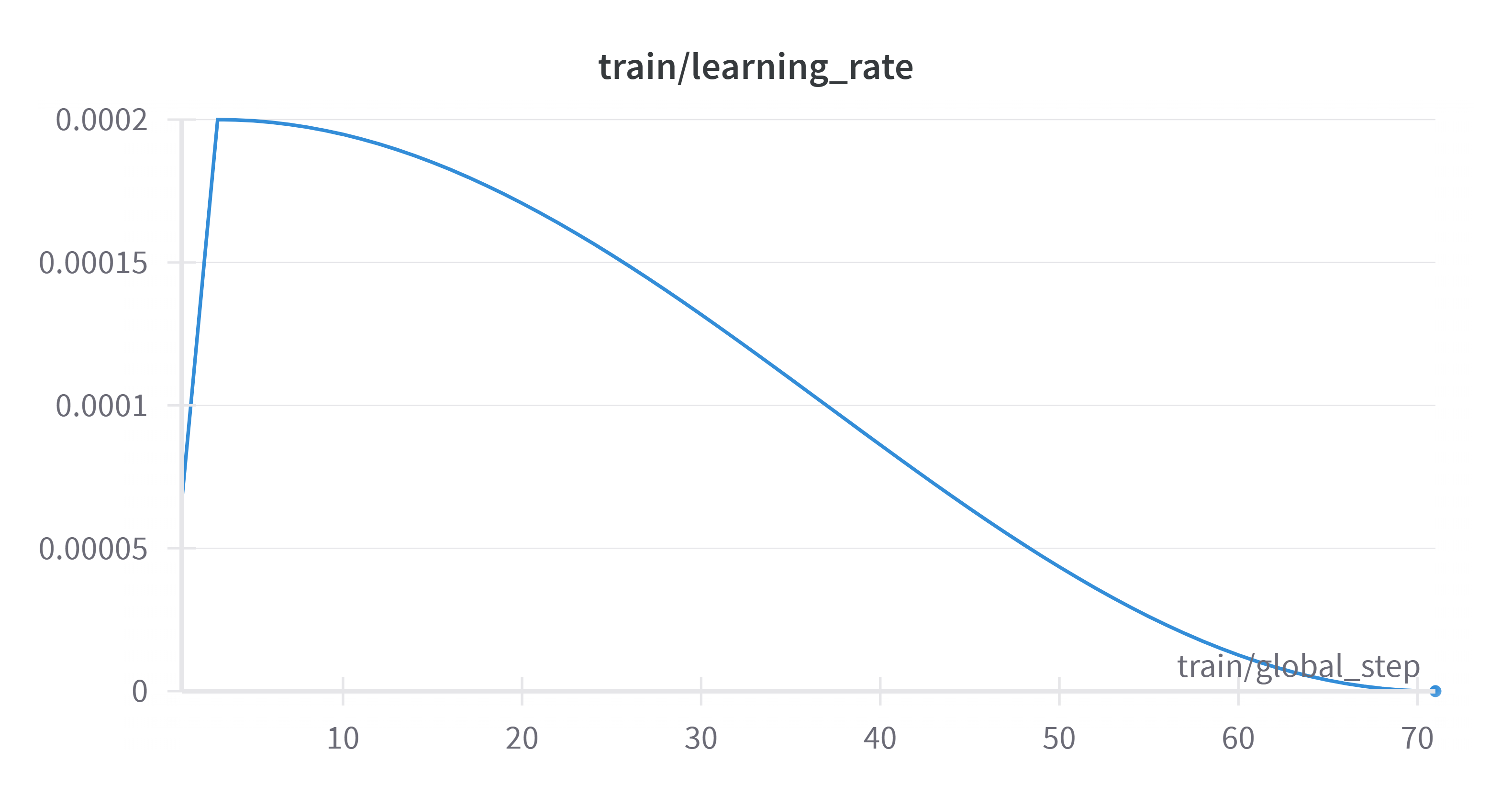}
        \caption{Learning Rate}
    \end{subfigure}
    \hfill
    \begin{subfigure}[b]{0.49\textwidth}
        \includegraphics[width=\textwidth]{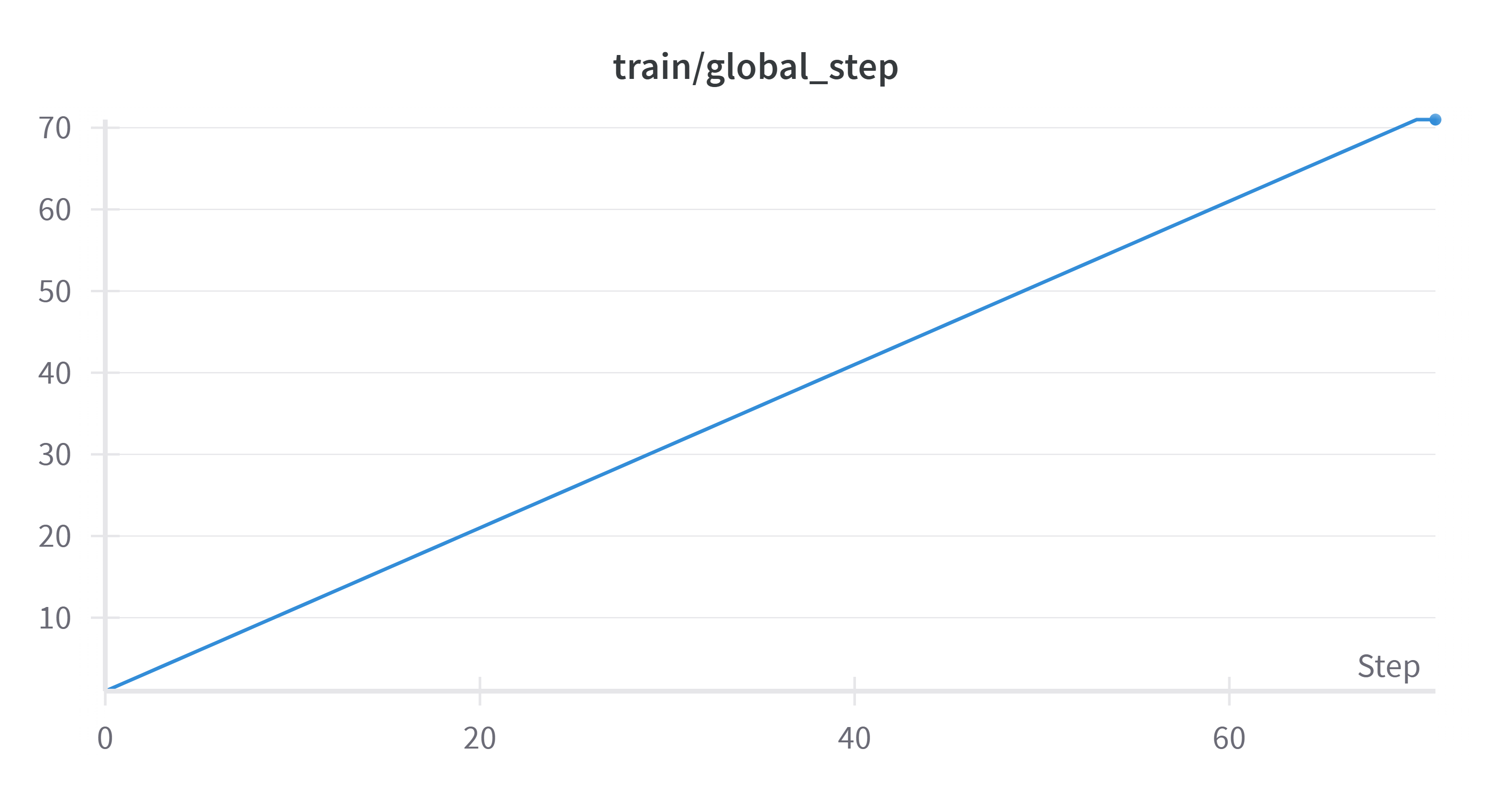}
        \caption{Global Step}
    \end{subfigure}
    \hfill
    \begin{subfigure}[b]{0.49\textwidth}
        \includegraphics[width=\textwidth]{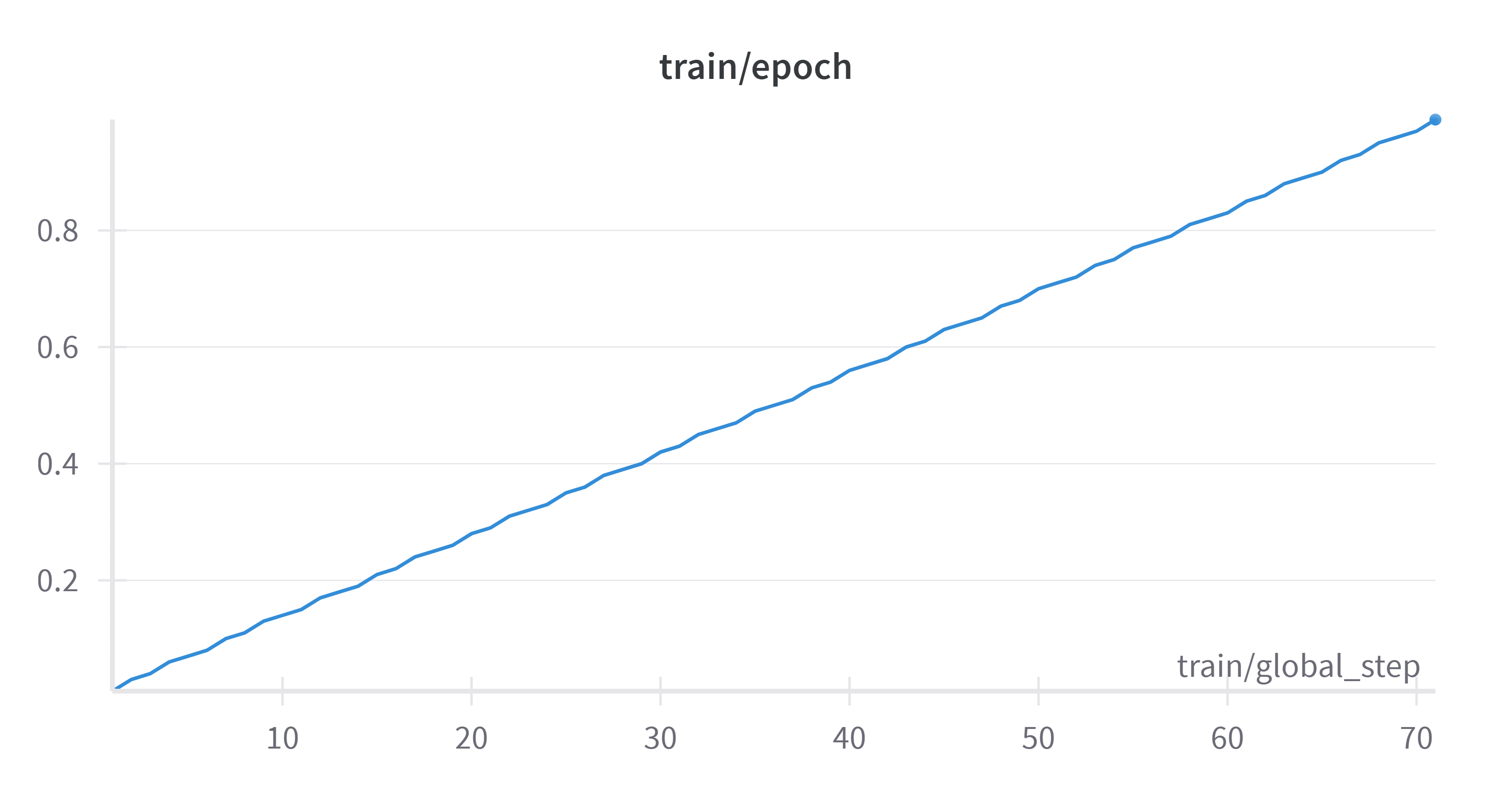}
        \caption{Epochs}
    \end{subfigure}
    \vspace{2em}
    \caption{\textbf{Tracking Data from WandB (Training).} (a) Depicts the decay of the training loss over steps, indicating effective learning and convergence of the model. (b) Illustrates a learning rate schedule where the rate initially increases slightly before gradually decreasing, following a cosine strategy. (c) Shows a linear increase in the global training steps. (d) Displays the linear progression of training epochs with respect to global steps.}
    \label{fig:seven_combined1}
\end{figure*}

Figure \ref{fig:seven_combined1} illustrates the Weights \& Biases (WandB) tracking data from the final iteration of LoRA fine-tuning applied to LLaVA-v1.5-7B within the proposed STL framework, specifically targeting the language science domain. 

\begin{figure*}
  \centering
  \includegraphics[width=0.95\textwidth]{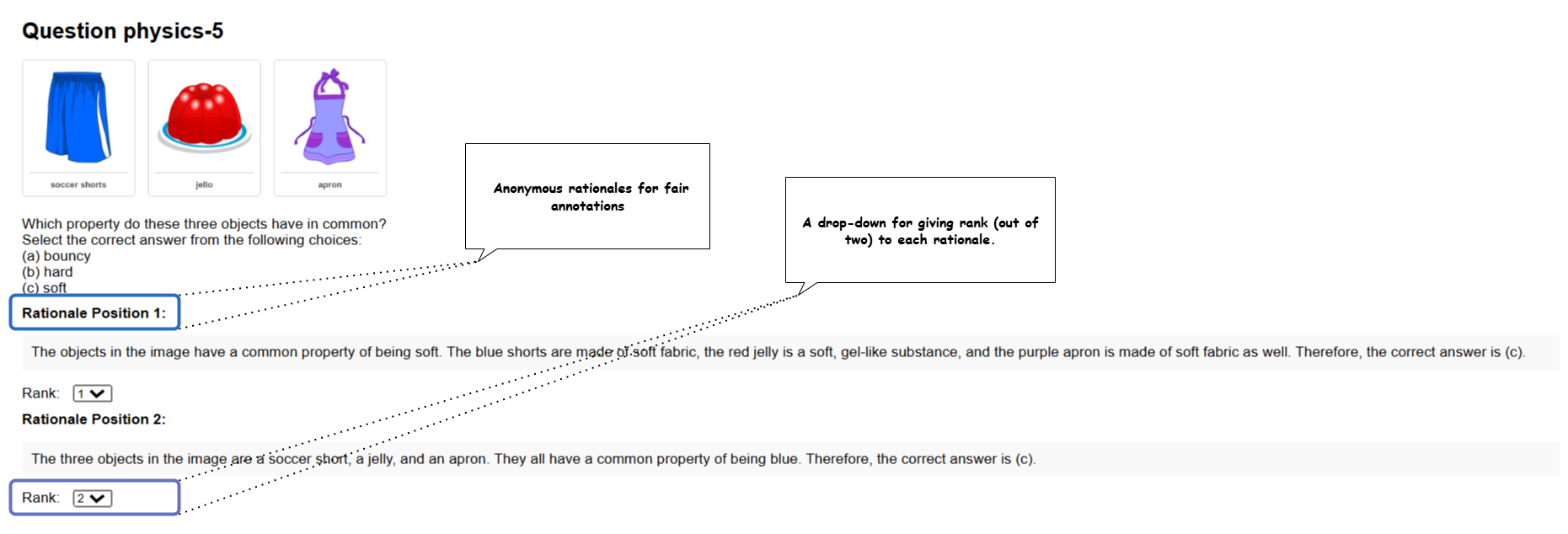}
  \caption{\textbf{Screenshot of our subjective annotation GUI.} Please note that the name of the rationale generator model has been kept anonymous to ensure a fair comparison.}
  \label{fig:subjective_gui}
\end{figure*}

\section{Subjective Analysis}

Figure \ref{fig:subjective_gui} illustrates the interface developed to collect these annotations.


\end{document}